\documentclass[mlabstract]{jmlr}

\usepackage{longtable}

\usepackage{booktabs}
\usepackage[load-configurations=version-1]{siunitx} 

\theorembodyfont{\upshape}
\theoremheaderfont{\scshape}
\theorempostheader{:}
\theoremsep{\newline}

\jmlrvolume{}
\firstpageno{1}
\editors{Sophia Sanborn, Christian Shewmake, Simone Azeglio, Nina Miolane}

\jmlryear{2023}
\jmlrworkshop{Symmetry and Geometry in Neural Representations}

\usepackage[utf8]{inputenc} 
\usepackage[T1]{fontenc}    
\usepackage{hyperref}       
\usepackage{url}            
\usepackage{booktabs}       
\usepackage{amsfonts}       
\usepackage{nicefrac}       
\usepackage{microtype}      
\usepackage{xcolor}         

\usepackage[disable]{todonotes}

\title[Algebraic Topological Networks]{Algebraic Topological Networks \\ via the Persistent Local Homology Sheaf}

\author{
  \Name{Gabriele Cesa} \Email{gcesa@qti.qualcomm.com}\\
  \Name{Arash Behboodi} \Email{behboodi@qti.qualcomm.com} \\
  \addr{
    Qualcomm AI Research, Amsterdam\nametag{\thanks{Qualcomm AI Research is an initiative of Qualcomm Technologies, Inc.}}
  }
}

\usepackage{amsmath,amsfonts,bm}
\usepackage{mathtools}

\usepackage{quiver} 

\def\1{\bm{1}}

\def\vv{{\bm{v}}}

\def\vx{{\bm{x}}}

\DeclareMathAlphabet{\mathsfit}{\encodingdefault}{\sfdefault}{m}{sl}
\SetMathAlphabet{\mathsfit}{bold}{\encodingdefault}{\sfdefault}{bx}{n}

\def\gB{{\mathcal{B}}}

\def\gF{{\mathcal{F}}}

\def\gH{{\mathcal{H}}}

\newcommand{\R}{\mathbb{R}}
\newcommand{\Nat}{\mathbb{N}}

\newcommand{\Zp}{\mathbb{Z}/p\mathbb{Z}}

\newcommand{\F}{\mathbb{F}}

\DeclareMathOperator{\ima}{im}

\DeclareMathOperator{\coima}{coim}

\renewcommand{\S}{\textnormal{S}} 

\DeclareMathOperator{\st}{star}
\DeclareMathOperator{\cl}{cl}
\DeclareMathOperator{\ins}{int}

\begin{document}

\maketitle


\begin{abstract}
In this work, we introduce a novel approach based on algebraic topology to enhance graph convolution and attention modules by incorporating local topological properties of the data.
To do so, we consider the framework of \emph{sheaf neural networks}, which has been previously leveraged to incorporate additional structure into graph neural networks' features and construct more expressive, non-isotropic messages.
Specifically, given an input simplicial complex (e.g. generated by the cliques of a graph or the neighbors in a point cloud), we construct its \emph{local homology sheaf}, which assigns to each node the vector space of its local homology.
The intermediate features of our networks live in these vector spaces and we leverage the associated sheaf Laplacian to construct more complex linear messages between them.
Moreover, we extend this approach by considering the \emph{persistent} version of local homology associated with a weighted simplicial complex (e.g., built from pairwise distances of nodes embeddings).
This \emph{i)} solves the problem of the lack of a natural choice of basis for the local homology vector spaces and \emph{ii)} makes the sheaf itself differentiable, which enables our models to directly optimize the topology of their intermediate features.
\end{abstract}



\section{Introduction}

Many works in the literature extended standard \textit{Graph Convolution Networks} (GCNs) \cite{kipf2016semi}, which rely on isotropic message passing along a graph's edges, to more expressive message passing operators.
\textit{Sheaf neural networks} \cite{hansen2020sheaf} provide a generic framework to encode more structure into the features attached to a graph's nodes, which can be leveraged to define more expressive messages between the feature spaces of neighboring nodes via the sheaf's restriction maps and the \emph{sheaf Laplacian}.
Briefly, a \textbf{sheaf} $\mathcal{F}$ on a space $X$ associates a (feature) vector space $\mathcal{F}(U)$ to each (open) set $U \subset X$ and a linear map $\mathcal{F}(U \subset V)$ to each pair $U \subset V$, i.e. the \emph{restriction map}.
Two restrictions $\mathcal{F}(W \subset U)^T\mathcal{F}(W\subset V)$ can be combined to send messages between $U$ and $V$ via their intersection $W = U \cap V$: this is the idea behind the \emph{sheaf Laplacian}.
While a sheaf should also satisfy \emph{locality} and \emph{gluing properties}, these are not necessary to construct the Laplacian and are usually ignored in neural networks; see Apx.~\ref{sec:sheaves} for more details.
In practice, sheaf neural networks associate a feature vector space to each node in a graph and a linear map to each edge, relating the feature spaces of connected nodes.
With respect to the graph Laplacian, this new Laplacian doesn't enforce similarity between neighboring nodes' features, thereby circumventing the homophily assumption \cite{bodnar_neural_2022}.

GCNs are the simplest example of sheaf neural networks: these architectures rely on a sheaf which associates the same vector space to each node and whose restriction maps are identities.
This enables a simple weight sharing at the cost of less expressive message passing.
Other works can be interpreted under this lens: \cite{de2020natural} constructs a very expressive sheaf over graphs where each node has a feature dimension for each of its neighbors and restriction maps match dimensions corresponding to the same nodes\footnote{Messages are actually constructed with something more similar to a cosheaf Laplacian by leveraging the union rather than the intersection of open sets. The work also supports more generic feature spaces.}.
Alternatively, since datasets rarely come with a sheaf structure already defined, \cite{bodnar_neural_2022} propose learning to predict restriction maps from input features during inference.

\paragraph{Contributions}
We use tools from \textit{algebraic topology} \cite{hatcher2002algebraic} to construct a new sheaf for neural networks: the \textbf{Local Homology} sheaf in the \textit{flag complex} of a graph \cite{robinson2018local}.
This sheaf catches local topological features of a space: it associates to each node a feature vector space with a component for each "relative cycle" in its neighborhood.
Intuitively, an order $k$ local relative cycle detects a subspace which locally looks like a $k$-dimensional manifold.
For this reason, the local homology sheaf is typically used for \textbf{stratification detection} of triangulated spaces.
Interestingly, sheaf diffusion along the edges is sufficient to detect higher order (local and global) homological properties of the space, with no need of higher-order simplicial message passing.

Unfortunately, the homology sheaf doesn't prescribe a natural choice of basis for the feature vector space, which makes constructing learnable linear and activation layers challenging.
We tackle this limitation by considering \emph{weighted graphs} and leveraging \textbf{persistent homology}, the standard tool in \emph{Topologial Data Analysis} \cite{carlsson2009topology}.
Finally, this new construction generates a sheaf whose Laplacian is \emph{differentiable} with respect to the graph weights, which can be output of another learnable module (e.g. from learnable node embeddings): this enables our model to learn the sheaf structure or tune the weights in a topological~informed~way.


\section{Simplicial Complexes, Homology and the Local Homology Sheaf}
\label{sec:local_homology_sheaf}

We first briefly review some essential concepts but see Apx.~\ref{apx:simplicial_homology} for more details.

\textbf{Simplicial Complexes }
Assume a \emph{finite} set $V$ of $|V|=N$ nodes.
A \textbf{simplicial complex} is a collection $S \subset 2^V$ of subsets of $V$; a subset $\sigma \in S$ with $k+1$ elements is called a $k$-\textbf{simplex}.
Simplicial complexes generalize the common notion of \emph{graph} beyond pairwise relationships.
For example, if $G=(V, E)$ is a graph, its \textbf{flag} (or \textbf{clique}) \textbf{complex} is a simplicial complex $S$ with nodes $V$ and containing a simplex for each \emph{clique} in $G$, i.e. for each set of nodes in $G$ which form a complete subgraph.

\textbf{Chains and Boundaries }
The graph Laplacian can be constructed from the \emph{incidence matrix} $\partial \in \R^{|V| \times |E|}$ as $\Delta_0 = \partial \partial^T$.
This construction generalizes to simplicial complexes.
A \textbf{$k$-chain} of $S$ is a scalar signal over (oriented) $k$-simplicies; $C_k(S)$, or just $C_k$, is the vector space of all $k$-chains.
The incidence matrix is generalized by the \textbf{boundary operator} $\partial_k\!:\!C_k\!\to\!C_{k-1}$, which models the relationship between each $k$-simplex and its \emph{faces} (its $k$-dimensional subsets).
The $k$-th \emph{Hodge Laplacian} is defined as $\Delta_k\!:=\!\partial_k^T\partial_k + \partial_{k+1}\partial_{k+1}^T\!:\!C_k\!\to\!C_k$ and has been used to construct a variety of simplicial neural networks \cite{papillon2023architectures}.

\textbf{Cycles and Homology }
A classical result in topology is that \emph{a boundary of a space has no boundary}: $\ima \partial_{k+1} \subset \ker \partial_k$.
The \textbf{$k$-th homology group} is the \emph{quotient vector space} $H_k(S) := \ker \partial_k / \ima \partial_{k+1}$.
Its dimensionality $\dim H_k$ is an important invariant counting the $k$-dimensional holes in $S$ and its basis can be thought as a set~of~independent~$k$-dimensional cycles in $S$ ($0$-cycles are connected components, $1$-cycles are loops, $2$-cycles are cavities).

\begin{figure}[t]
\floatconts
  {fig:relative_homology}
  {\caption{Examples of homology and relative homology. The {\color{gray} greyed out simplices} can be thought as being "collapsed" in a single point to compute relative homology: then, the {\color{blue}blue area $\beta$} turns into a $2$-sphere while the {\color{red} red line $\gamma$} turns into a $1$d ring.}}
  {\setlength{\jmlrminsubcaptionwidth}{0.28\linewidth}%
    \subfigure[$H_1(S)$]{\label{fig:relative_homology_1} \includegraphics[width=0.27\linewidth]{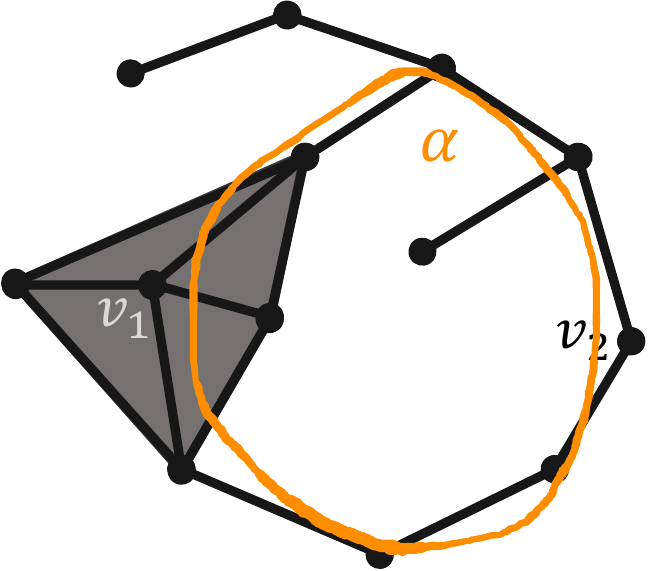}}%
    \hfill%
    \subfigure[$H_2(S, {\color{gray}S\ \backslash \st v_1})$]{\label{fig:relative_homology_rel_2} \includegraphics[width=0.27\linewidth]{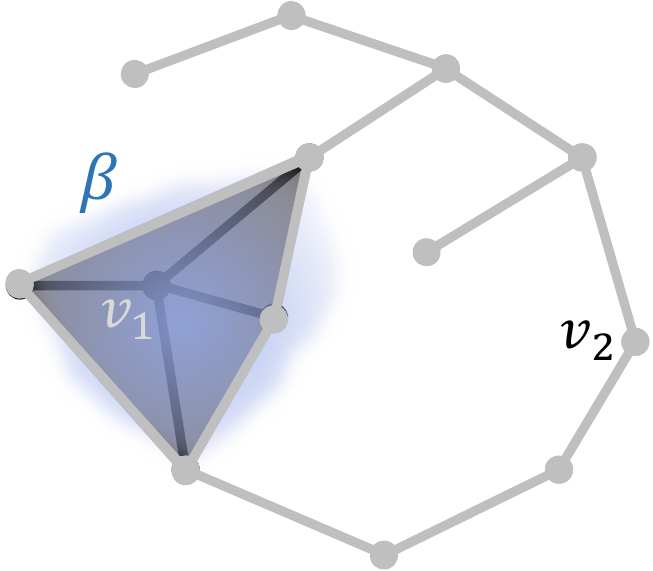}}%
    \hfill%
    \subfigure[$H_1(S, {\color{gray}S\ \backslash \st v_2)}$]{\label{fig:relative_homology_rel_1} \includegraphics[width=0.27\linewidth]{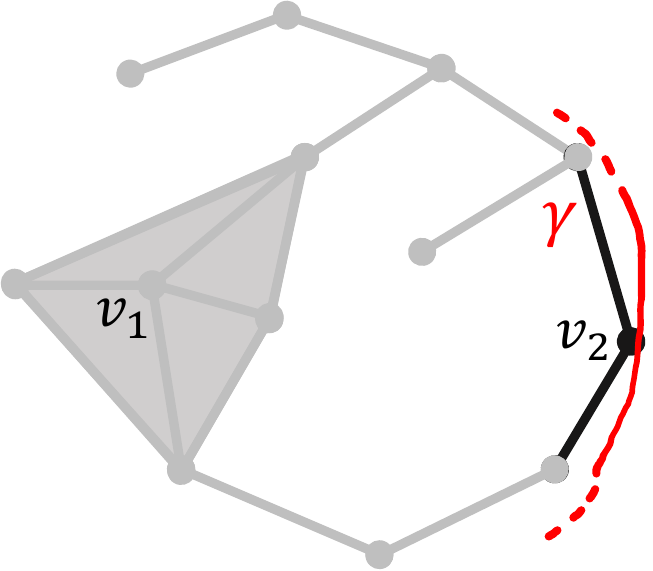}}%
  }
\end{figure}

Our construction is similar to \citep{robinson2018local}, which first introduced the \emph{Local Homology Sheaf} over simplicial complexes.
Given a $k$-simplex $\sigma \in S$, define its \textbf{star} as $\st \sigma = \{\tau \in S : \sigma \subset \tau\}$.
An \textit{open subset} $A \subseteq S$ is the union of sets of the form $\st \sigma$; note that this is not necessarily a simplicial complex.
Instead, a subset $A \subseteq S$ is \textit{closed} if it is a \textit{subcomplex} of $S$ (the faces of every simplex in $A$ are also in $A$).
We also define the \textbf{closure} $\cl A$ as the smallest subcomplex of $S$ containing $A$,
the \textbf{interior} $\ins A$ as the largest open set contained in $A$ and
the \textbf{frontier} as $\partial A = \cl A \ \backslash \ A$.

\textbf{Relative Homology}
Let $A \subseteq S$ be a \textit{subcomplex} of $S$.
The $k$-th \textbf{relative homology} $H_k(S, A)$ describes the $k$-th homology of the \emph{quotient space} $S / A$ obtained from $S$ by identifying all its points within $A$, i.e. by "collapsing" all points in $A$ in a single point.
Fig.~\ref{fig:relative_homology} shows a few examples.
However, note that the relative homologies $H_k(S, {\color{gray}S\ \backslash \st v})$ doesn't depend on {\color{gray}(most) gray simplices in $S \backslash \st v$}, but only on those in $\st v$ and its closest neighbors.
This is the \textbf{Excision Principle}: if $A \subset B \subset S$ are subsets of $S$ such that $\cl A \subset \ins B$, then $H_k(S,B) \cong H_k(S\backslash A, B\backslash A)$.
When $A \subset S$ is an open set, $H_k(S, S \backslash A) \cong H_k(\cl A, \partial A)$.

\textbf{Local Homology Sheaf}
As in \cite{robinson2018local}, we consider the sheaf $\mathcal{H}_*$ defined as $\mathcal{H}_*(A) = H_*(S, S\backslash A) \cong H_*(\cl A, \partial A)$ for each open set $A \subset S$ ($S \backslash A$ is closed if $A$ is open).
The sheaf structure is naturally given by the following \emph{long exact sequence}\footnote{
An \emph{exact sequence} is a sequence of maps s.t. the image of a map equals to the kernel of the~consecutive~one.
}:
\begin{equation}\label{eq:local_homology_sheaf_sequence}
\begin{tikzcd}
	{\cdots } & {\mathcal{H}_k(A \cup B)} & {\mathcal{H}_k(A) \oplus \mathcal{H}_k(B)} & {\mathcal{H}_{k}(A \cap B)} & \cdots
	\arrow["{k_* - l_*}", from=1-3, to=1-4]
	\arrow["{i_*,j_*}", from=1-2, to=1-3]
	\arrow[from=1-1, to=1-2]
	\arrow[from=1-4, to=1-5]
\end{tikzcd}
\end{equation}
where $k_*, l_*, i_*$ and $j_*$ are the sheaf restriction maps. 
This is a special case of the well known \textit{Mayer-Vietoris sequence}; see Apx.~\ref{apx:les}.
In particular, $\mathcal{H}_*(\st v_i)$ is called the \textbf{local homology} of the vertex $v_i$.
Intuitively, the local homology of a point in a topological space contains information about what the space looks like around that point.
If the space is an $n$-manifold, the local neighborhood $U$ of any point looks like a $n$-ball,  whose boundary $\partial U$ is isomorphic to a $n-1$-sphere $\mathcal{S}^{n-1}$.
Then, like in Fig.~\ref{fig:relative_homology}, via excision the local homology is $H_*(U, \partial U) \cong \tilde{H}_*(\S^n)$, i.e. the (reduced) homology of an $n$-sphere, which only has one cycle of order $n$.
Hence, local homology \emph{detects the local dimensionality of a space}.
Moreover, points at the boundary of the space have empty local homology.
This idea was used in \cite{robinson2018local}, among others, for \emph{stratification detection}.
Finally, note that the restriction maps constructed in Eq.~\ref{eq:local_homology_sheaf_sequence} are identity maps on $\mathcal{H}_n$ for points in the interior of an $n$-manifold\footnote{The local homology sheaf $\mathcal{H}_n$ is closely related to the \emph{orientation sheaf} of an $n$-manifold.}.

Finally, recall that \textbf{sheaf diffusion} minimizes the \emph{sheaf Dirichlet energy} of a signal \cite{bodnar_neural_2022}.
At zero energy, the signal is in the Laplacian's kernel and, by the sheaf property, belongs to the \textbf{global sections} of $\mathcal{H}(S)$  \cite{hansen2021opinion}.
Because $\mathcal{H}_k(S) = H_k(S, \emptyset) = H_k(S)$ (Corollary 20 \cite{robinson2018local}), diffusion converges towards the global homology classes of $S$ of any order $k$ while only relying on messages along edges.

\textbf{Persistent Homology}
provides a richer structure than homology, by enriching homology classes with a (differentiable) notion of resolution; see Apx.~\ref{apx:persistent_homology}.
Rather than building a single sheaf for a fixed complex $S$, we consider a \emph{filtration}, i.e. a sequence of simplicial complexes $\{S_t\}_t$ related by inclusion, and build the local homology sheaf of the complex $S_t$ at each time-step $t$.
Cycles in the local homology at a step in the filtration can "\emph{persist}" in the consecutive steps or disappear.
This enriches the local homology with a notion of time or scale, i.e. each cycle is associated with a time-step where it emerges and a time-step where it disappears.
In practice, we define the "filtered" neighborhood of a node $i$ as $A_i^t = \st^t v_i \subset S_t$ and 
\emph{compute the persistent cycles} in 
the \textbf{persistent module} $\mathcal{H}_k^\bullet(A_i) = \bigoplus_t H_k(S^t, S^t \backslash A^t_i)$
as in Apx.~\ref{subsec:computing_relative_homology}.
Persistent cycles are shared among the time-steps between their births and deaths, see Eq.~\ref{eq:persistent_relative_homology}.
This \emph{feature sharing strategy} generates the \textbf{persistent relative homology} subspace $\mathcal{H}_k(A_i) \subset \mathcal{H}^\bullet_k(A_i)$.
Columns in Fig.~\ref{fig:persistent_laplacian} are examples of persistent local homology.

\section{Proposed Architecture}
\label{sec:method}

Given a graph $G=(V, E)$ with weighted edges (e.g. the distance matrix of a point cloud), we construct the \emph{Vietoris-Rips filtration}\footnote{
    A simplex appears in the filtration at a time step equal to the maximum weight of its edges.
}
$\{S_t\}_{t}$
of its flag complex $S$.
Unfortunately, while the persistent module $\mathcal{H}^\bullet_k$ forms a sheaf, persistent local homology $\mathcal{H}_k\subset \mathcal{H}^\bullet_k$ fails to be a sheaf \cite{palser2019excision}.
To preserve the sheaf diffusion properties described before, we prefer using the sheaf Laplacian of $\mathcal{H}^\bullet_k$.
Hence, our \textbf{message passing} on $\mathcal{H}_k$ first \emph{embeds} persistent homology features in the sheaf $\mathcal{H}_k^\bullet$, then \emph{applies} the sheaf Laplacian $\Delta_{\mathcal{H}_k^\bullet}$ and, finally, \emph{projects} the output on $\mathcal{H}_k$ by averaging the features of a cycle along its life span.
Fig.~\ref{fig:persistent_laplacian} shows an example of Laplacian $\Delta_{\mathcal{H}_k^\bullet}$.
See Apx.~\ref{apx:sheaf_laplacian} for details on the implementation.

To complete our architecture, we need to include a \textbf{learnable layer} operating on each node's feature space $\mathcal{H}_*(A_i)$.
This involves two challenges: \emph{i)} a persistent cycle is only \emph{defined up to a sign} (the Laplacian constructed is equivariant to these sign changes) and \emph{ii)} each node's feature space looks different.
\emph{i)} is related to the spectral symmetries studied in \cite{lim2023expressive} and be can solved similarly: given $\vx \in \mathcal{H}_*(A_i)$, we construct a sign equivariant layer of the form $\psi(\vx) = \vx \circ \rho(|\vx|)$.
The learnable operator $\rho$ can be modeled by a simple MLP.
To share $\rho$ among different nodes and solve \emph{ii)}, we learn a separate MLP $\Psi$ to output the weights of $\rho$ for each node individually.
Note that each persistent cycle is uniquely identified by its order $k$ and its birth and death times $s, t \in \R$, 
Then, we can parameterize a linear map on $\mathcal{H}_*(A_i)$ via $\Psi$ as follows: for each pair $(i, j)$ of input/output persistent cycles, the $(i, j)$-th entry of the weight matrix is parameterized by $\Psi(k_i, s_i, t_i, k_j, s_j, t_j) \in \R$.
As usual, this approach can be integrated in a multi-channel network, where the features of the node include multiple copies of the vector space $\mathcal{H}_*(A)$.

\section{Limitations and Complexity}
Persistent homology is computed by reducing the boundary matrices, with a worst case complexity cubic in the number of simplices.
Assuming $N$ nodes and by considering only homology up to order $K$ (typically $K=2$ or $3$), there are at worst $O(N^{K+1})$ simplices so the complexity is $O(N^{3K+3})$.
However, thanks to the excision principle, local homology can be computed by using only a limited number of neighboring nodes.
Assuming each node has $O(n)$ neighbors, computing the local homology of each node costs only $O(Nn^{3K+3})$.
Moreover, the computation of each local homology can be fully parallelized.
For each par of nodes, the sheaf Laplacian is also computed via a matrix reduction using the union of their local neighbors (with $O(2n)$ nodes): with a similar worst case complexity $O((2n)^{3K+3})$ for each pair of nodes $O(nN)$, the overall complexity is then $O(N n^{3K+4} 2^{3K+3})$.
Still, we note that there exists optimized algorithms like \texttt{Ripser} \cite{bauer2021ripser}, which are much faster on average by leveraging a number of smart heuristics; see also \cite{BAUER201776} for a more detailed discussion.
Additionally, the number of neighbors $n$ can be chosen sufficiently low to control the overall complexity.
The main limitation we currently see is the fact that these computations can not be performed on a GPU in a straightforward way.
As a result, computing the sheaf structure requires moving the edge weights to the CPU during inference and, then, move the sheaf Laplacian data back on GPU.

\section{Conclusions and Discussions}
The proposed local homology sheaf Laplacian can be used to enhance existing deep learning architectures by making them aware of the local and global topology of the underlying data structure during inference.
Previous works \citep{pmlr-v97-rieck19a,hofer2020graph,carriere2020perslay,horn2021topological}
already successfully augmented graph neural networks with global topological features by leveraging the persistent homology of weighted input graphs.
The proposed local homology sheaf can be used in a similar way to enrich each node in a graph with its local topological features while the sheaf Laplacian relates algebraically these local features.
As argued at the end of Sec.~\ref{sec:local_homology_sheaf}, global topological features are instead encoded in the global sections of this sheaf, i.e. the kernel of the proposed Laplacian.
We expect this to be especially useful in tasks such as graph link prediction, mesh reconstruction or simply where the data presents a variety of topologies.
We plan to experimentally evaluate this method on similar tasks in future works.

\acks{
  We thank Giovanni Luca Marchetti for the very insightful discussions about efficiently computing the sheaf Laplacian, the Mayer-Vietoris sequences and~other~algebraic~topology~ideas.
}

\newpage

\bibliography{bibliography}

\newpage

\appendix

\section{Example of persistent sheaf Laplacian}
\begin{figure}[ht]
    \centering
    \includegraphics[width=0.77\linewidth]{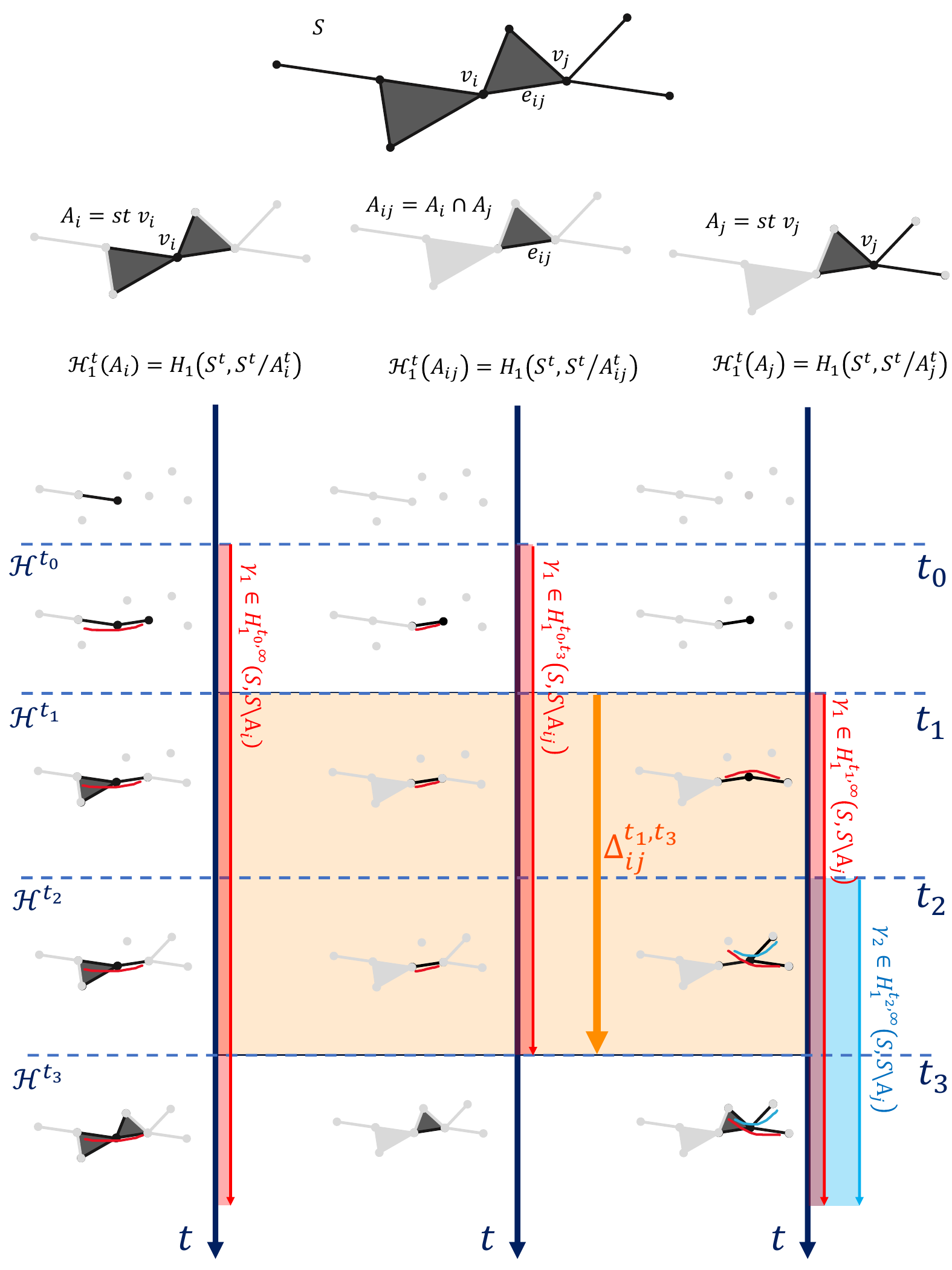}
    \caption{\small 
      Example of persistent sheaf Laplacian.
      The three columns depict the time evolution of the filtrations of the local neighborhood of three simplices. 
      At different time steps, some new relative cycles appear or disappear and each cycle "persists" for an interval of time.
      In our architecture, a single feature is stored for each persistent cycle; this feature can be thought as been shared over all time steps within the cycle's life span.
      Moreover, the three columns share a {\color{red} relative $1$-cycle $\gamma_1$}.
      Note that this cycle exists at different times intervals in the three columns and, therefore, there exists a sheaf Laplacian only during the {\color{orange} intersection of these intervals $[t_1, t_3)$}.
    }
    \label{fig:persistent_laplacian}
\end{figure}

\newpage


\section{Sheaves}
\label{sec:sheaves}

Given a space $X$, a \textbf{pre-sheaf} $\mathcal{F}$ associates to each open set $U \subset X$ a space $\mathcal{F}(U)$ and to each pair $U \subset V (\subset X)$ a map $\gF(U \subset V): \mathcal{F}(V) \to \mathcal{F}(U)$ (\emph{restriction map}), such that $\gF(U \subset U)$ is the identity and $\gF(U \subset V) \gF(V \subset W) = \gF(U \subset W)$ if $U \subset V \subset W$.
We are mostly interested in the cases where $\mathcal{F}(U)$ are vector spaces.
An element of $\mathcal{F}(U)$ is called a "\textbf{local section}", while an element of $\mathcal{F}(X)$ is called a "\textbf{global section}".

Given an open cover $\{U_i \subset U\}_i$ of $U \subset X$, a \textbf{sheaf} is a pre-sheaf satisfying two additional axioms: 
\begin{enumerate}
    \item \emph{locality}: if two local sections $s, t \in \mathcal{F}(U)$ agree when restricted on all $\{U_i \}_i$, then they are identical
    \item \emph{gluing}: if a set of local sections $\{s_i \in \mathcal{F}(U_i)\}_i$ agree on all their overlaps, then there exists section $s \in \mathcal{F}(U)$ which agrees with $s_i$ when restricted on $U_i$, for all $i$
\end{enumerate}

Given a sheaf $\mathcal{F}$, with $\mathcal{F}(U)$ vector spaces, we can construct the \emph{sheaf Laplacian} \cite{hansen2020sheaf}.
To do so, consider an open cover $\{U_i\}_i$ of the space $X$.
For any $i, j$ s.t. $U_i \cap U_j \neq \emptyset$, define the linear map 
\[
    \delta_{ij}: \mathcal{F}(U_i) \times \mathcal{F}(U_j) \to \mathcal{F}(U_i \cap U_j), \quad x_i, x_j \mapsto \gF(U_i\cap U_j \subset U_i) x_i - \gF(U_i \cap U_j \subset U_j) x_j
\]
Then, the sheaf Laplacian is a block matrix defined as $L_\mathcal{F} = \delta^T \delta$.
Note that, if $i\neq j$, the $(i, j)$-th block is defined as $[\Delta_\mathcal{F}]_{ij} = -\gF(U_i \cap U_j \subset U_i)^T\gF(U_i \cap U_j \subset U_j)$.

Given a sheaf defined over a graph, the sheaf Laplacian generalizes the classical graph Laplacian and provides a useful tool to build more expressive message passing operators for neural networks.
\todo{cite}

To build message passing, the restriction map of a pre-sheaf is sufficient and we do not actually need the additional two axioms of a sheaf.
Still, since local homology in Sec.~\ref{sec:local_homology_sheaf} forms a sheaf with some interesting properties and to keep the notation simpler, we use the word "sheaf" also in the message passing architectures which don't enforce these axioms.

\section{Simplicial Complexes, Boundary Maps and Homology}
\label{apx:simplicial_homology}

\paragraph{Simplicial Complex} 
Give a \emph{finite} set of nodes $V$, with $|V|=N \in \Nat$, a \textbf{simplicial complex} $S$ is a mathematical objects that can be thoughts as a collection of subsets of $V$, i.e. $S \subset 2^V$, such that $\forall \sigma \in S, \tau \subset \sigma \implies \tau \in S$.
Each such subset $\sigma \in S$ is called a \textbf{simplex}.
We usually refer to simplices with $k+1$ elements as $k$-simplices.
Simplicial complexes generalize the common notion of \emph{graph}, by thinking of an edge as a set containing two nodes.
A $k$-simplex $\sigma = \{v_0, \dots, v_k\} \subset V$, like edges, is typically associated with an \textbf{orientation}, i.e. a particular choice of ordering of its elements $\sigma = [v_0, \dots, v_k] \in S$.
Two $k$-simplicies $\sigma, \sigma'$ containing the same subset of nodes share the same orientation if they differ by an \emph{even permutation} but have opposite orientation if they differ by an \emph{odd permutation}.

\paragraph{Chain Complexes and Boundary Operators}
Let $S$ be a simplicial complex. 
A \textbf{$k$-chain} of $S$ is a scalar function $f: S \to \R$ on the oriented $k$-simplicies of $S$ such that $f(\sigma) = f(\sigma')$ if $\sigma$ and $\sigma'$ have the same orientation (differ by an even permutation) and $f(\sigma) = -f(\sigma')$ if they have opposite orientation (differ by an odd permutation).
A \textbf{chain complex} $C_\bullet(S)$ is a sequence of vector spaces $C_0(S), C_1(S), \dots$, where $C_k(S)$ is the vector space of all $k$-chains.
A chain complex is associated with a \emph{linear} \textbf{boundary operator} (or \emph{differential}) $\partial_k: C_k \to C_{k-1}$, defined on a $k$-simplex $\sigma=[v_0, v_1, \dots, v_k] \in S$ (intended as one of the basis elements of $C_k$) as\footnote{This definition should be extended linearly to the full space $C_k$.}
\begin{align}
    \partial_k \sigma := \sum_{i=0}^k (-1)^i [v_0, \dots, \hat{v_i}, \dots, v_k]
\end{align}
where $[v_0, \dots, \hat{v_i}, \dots, v_k]$ is a $k-1$-simplex obtained from $\sigma$ by removing the node $v_i$.
We often use $\partial_\bullet: C_\bullet \to C_\bullet$ to denote the operator acting on each subspace $C_k$ of $C_\bullet$ with the corresponding operator $\partial_k$.

\paragraph{Example}
If $S$ is just a graph $G=(V, E)$, $C_0$ are functions over the nodes $V$ while $C_1$ are functions over the (oriented) edges $E$.
Moreover, the operator $\delta_1: C_1 \to C_0$ maps an edge $e = (v_0, v_1) \in E$ to $\partial_1(e) = [v_1] - [v_0]$ and, therefore, if $f \in C_1$, then
\begin{align}
    (\partial_1 f)(v_i) = \sum_{e=[v_j, v_i] \in E} f(e) - \sum_{e=[v_i, v_j] \in E} f(e)
\end{align}
This boundary operator $\partial_1$ can be used to construct the \emph{Graph Laplacian} as $\Delta_0 := \partial_1 \partial_1^T$, which is typically used to perform message passing in GCNs.
The boundary operators can be used to generalize this construction to a \emph{Hodge Laplacian} over a simplicial complex, defined as $\Delta_k := \partial_k^T\partial_k + \partial_{k+1}\partial_{k+1}^T : C_k \to C_k$, which can be used to construct a variety of higher-order simplicial neural networks \cite{papillon2023architectures}.

\paragraph{Cycles, Boundaries and Homology}
A $k$-chain is said to be a \textbf{boundary} if it is the boundary of a $k+1$-chain; the subspace of $k$-boundaries is indicated by $B_k := \ima \partial_{k+1}$.
A $k$-chain is said to be a \textbf{cycle} if its boundary is zero; the subspace of $k$-cycles is indicated by $Z_k := \ker \partial_k$.
A classical result in topology is that \emph{a boundary of a space has no boundary}, i.e. $\partial_\bullet \circ \partial_\bullet = \partial_\bullet^2 = 0$.
It follows that $B_k = \ima \partial_{k+1} \subset Z_k = \ker \partial_k$.
The \textbf{$k$-th homology group} is defined as the \emph{quotient vector space} $H_k(S) := Z_k / B_k$.
The dimensionality $\dim H_k$ is an important invariant and is equal to the $k-th$ \emph{Betti number} $\beta_k$ of $S$, which counts the $k$-dimensional holes in $S$.

\paragraph{Topology, open sets and subcomplexes of a simplicial complex}
Given a finite simplicial complex $S$, a subset $A \subseteq S$ is said to be \textit{closed} if it is also a simplicial complex (i.e. for each simplex in $A$, all its faces are also in $A$), i.e. it is a \textit{subcomplex} of $S$.
Instead, an \textit{open subset}\footnote{Formally, we consider the \textit{Alexandrov topology} of the simplicial complex like in \cite{robinson2018local}} $A \subseteq S$ is the union of sets of the form $\st \sigma = \{\tau \in S : \sigma \subset \tau\}$; note that this is not necessarily a simplicial complex.
Finally, we define a few useful operations on a subset $A \subseteq S$:
\begin{itemize}
    \item $S \backslash A$ indicates the standard set difference.
    \item the \textit{closure} $\cl A$ is the smallest subcomplex of $S$ containing $A$.
    \item the \textit{star} $\st A$ is the set of all simplices in $S$ which contain a simplex in $A$
    \item the \textit{boundary} (or frontier) $\partial A = \cl A \cap \cl (S \backslash A)$ 
    \item the \textit{interior} $\ins A$ is the largest open set contained in $A$ 
\end{itemize}

\paragraph{Relative Homology}
Let $A \subseteq S$ be a \textit{subcomplex} (i.e. a closed subset) of the simplicial complex $S$.
The \textit{relative $k$-chain space} $C_k(S, A) \cong C_k(S) / C_k(A)$ is the vector space of $k$-chains over $S$ which are zeros over the simplices in $A$.
Clearly, $C_k(S, A)$ is a subspace of $C_k(S)$ so the map $\partial_k: C_k \to C_{k-1}$ can be generalized to $\partial_k: C_k(S, A) \to C_{k-1}(S, A)$.
Then, the sub-space of relative $k$-boundaries is indicated by $B_k(S, A) := \ima \partial_{k+1}$ and the subspace of relative $k$-cycles is indicated by $Z_k(S, A) := \ker \partial_k$.
Finally, the $k$-th \textbf{relative homology} is defined as $H_k(S, A) = Z_k(S,A) / B_k(S, A)$.
Intuitively, $H_k(S, A)$ describes the $k$-th homology of the \textbf{quotient space} $S / A$ obtained from $S$ by identifying all its points within $A$, i.e. by "collapsing" all points in $A$ in a single point\footnote{Note the difference between the set difference $S\backslash A$ and the quotient space $S / A$.}.

\subsection{Properties of Homology and Long Exact Sequences}
\label{apx:les}

\paragraph{Long Exact Sequence for the Relative Homology}
If $A \subset S$ is a subcomplex of $S$, the relative chains give rise to a chain complex of relative homology groups with the following \textit{short exact sequence}:
\begin{align}
\label{eq:les_relative_homology}
    \dots \to H_k(A) \to^{i_k} H_k(S) \to^{j_k} H_k(S, A) \to^{\partial} H_{k-1}(A) \to \dots
\end{align}
The map $i_k$ comes from the inclusion of $C_k(A)$ into $C_k(S)$ and, intuitively, is relating the $k$-dimensional holes in $A$ with their copy in $S$.
The map $j_k$ comes from the projection of $C_k(S)$ into $C_k(S, A)$ and, intuitively, relates the holes in $S$ outside of $A$ with their copies in $S / A$.
Finally, the last map $\partial$ detects the $k$-dimensional holes in $S / A$, not present in $S$, which have appeared by collapsing $A$ in a single point.
These $k$-dimensional holes can be related with $A$'s $k-1$-dimensional boundary $\partial A \subset A$ and, therefore, included in $H_{k-1}(A)$.

\paragraph{Mayer-Vietoris Sequence}
Given two subcomplexes $A, B$ and the union $S = \ins A \cup \ins B$, there is another important \textit{long exact sequence}:
\begin{align}
\label{eq:les_mayervietoris}
    \dots \to H_{k+1}(S) \to^{\partial_*} H_k(A \cap B) \to^{i_*, j_*} H_k(A) \oplus H_k(B) \to^{k_* - l_*} H_k(S) \to \dots
\end{align}
Intuitively, if a $k+1$ cycle in $S$ is "broken" when $S$ is split into $A$ and $B$, the cycle splits into two $k+1$ chains in $A$ and $B$ which overlap in $A \cap B$. 
The boundaries of the two $k+1$ chains are homologous, i.e. they are a $k$-cycle in $A \cap B$, that is an element of $H_k(A\cap B)$.

This sequence holds also for relative homology, i.e. 
if $T = \ins C \cup \ins D \subset S$, with $C, D \subset S$, then
\begin{align}
\label{eq:les_mayervietoris_relative}
    \dots \to H_{k+1}(S, T) \to^{\partial_*} H_k(S, C \cap D) \to^{i_*, j_*} H_k(S, C) \oplus H_k(S, D) \to^{k_* - l_*} H_k(S, T) \to \dots
\end{align}

Eq.~\ref{eq:les_mayervietoris_relative} can also be used to construct the sequence in Eq.~\ref{eq:local_homology_sheaf_sequence} by replacing $C=S\backslash A, D=S \backslash B$ and, therefore, $T = C \cup D = S\backslash (A\cap B), C\cap D = S \backslash (A \cup B)$:
\begin{align}
    \dots \to H_k(S, S \backslash (A \cup B)) \to^{i_*, j_*} H_k(S, S\backslash A) \oplus H_k(S, S\backslash B) \to^{k_* - l_*} H_k(S, S \backslash (A \cap B)) \to \dots
\end{align}
Note that the maps in this sequence are given by the restriction maps of the sheaf and the exactness of the sequence proves exactly the gluing property of a sheaf. 
See Proposition 19 \cite{robinson2018local} for a more precise proof.

\section{Persistent Homology}
\label{apx:persistent_homology}

Given a finite simplicial complex $S$ and a function $f: S \to \R$ s.t. $f(\sigma) \leq f(\tau)$ if $\sigma < \tau$, define the simplicial complex $S_t = \{\sigma \in S : f(\sigma) \leq t\} \subset S$.
Note that $S_{t_1} \subset S_{t_2}$ if $t_1 \leq t_2$ and there exists $t^-, t^+$ such that $S_{t} = \emptyset$ for any $t \leq t^-$ and $S_t = S$ for any $t \geq t^+$.
Moreover, the sequence of simplicial complexes $\{S_t\}_{t \in \R}$ only contains a finite number of different complexes, so it can be replaced by a finite sequence $\{S_t\}_{t\in R}$ indexed by a subset $R \subset \R$.
This sequence is called a \textbf{filtration} of simplicial complexes.

The inclusion $S_{t_1} \subset S_{t_2}$ induces an homomorphism $i_k^{t_1, t_2}: H_k(S_{t_1}) \to H_k(S_{t_2})$, whose image $\ima i_k^{t_1, t_2}$ is the \textbf{persistent homology group} $H_k^{t_1, t_2}(S)$ and detects $k$-cycles in $S_{t_1}$ which are still present in $S_{t_2}$.
In particular, any $k$-cycles is born at a certain "time" $t_1$ (is not in the image of $i_k^{t, t_1}$ for any $t <t_1$).
It can also disappear at a time $t_2$ (it is in the kernel of $i_k^{t_1, t}$ for any $t \geq t_2$) or persist forever (it is a cycle in $H_k(S)$).

Note also that, if the complexes in the filtration only differ by a single simplex (i.e. the function $f$ gives a total ordering of the simplices), each time step a single $k$-simplex is added, which either creates a new $k$-cycle or destroys a $k-1$ cycle.
This is useful since the homology group $H_k(S)$ does not come with a natural choice of basis\footnote{
A basis for $H_k(S)$ can be computed as the $0$-eigenvectors of the $k$-th Hodge Laplacian $\Delta_k = \partial_k^T\partial_k + \partial_{k+1}\partial_{k+1}^T$. 
However, this basis is not unique and numerical algorithms are not guaranteed to return the same solution consistently.
The lack of a choice of basis is problematic to construct learnable neural operations like linear layers and non-linearities, which depend on a specific basis.
}; in this case, instead, cycles are uniquely identified by their birth and death times, which indirectly provides a 
choice of basis.

\textbf{Relative persistent homology} has also been studied in the literature, e.g. see \cite{robinson2018local,blaser2022relative}.
However, as far as we know, these works considered a slightly different formulation, assuming a filtration of pairs $(S, A_t)$, with $A_t \subset A_{t+1} \subset S$.

Instead, in this work, we consider a filtration of pairs in the following form.
Let $S^\infty = S$ and $A^\infty = A \subset S$.
Let $\mathbb{S} = (\dots, S_t, \dots, S^\infty=S)$ be a filtration of $S$ and $\mathbb{A} = (\dots, A_t, \dots, A^\infty=A)$ be a filtration of $A$, with $A_t = S_t \cap A$ (and, clearly, $S_t \subset S_{t+1}$ and $A_t \subset A_{t+1}$).
To simplify the notation, sometimes we just write $S$ instead of $\mathbb{S}$ to indicate a filtration.

As earlier, the inclusion $S_{t_1} \subset S_{t_2}$ induces an homomorphism $i_k^{t_1, t_2}: H_k(S_{t_1}, A_{t_1}) \to H_k(S_{t_2}, A_{t_2})$, whose image $\ima i_k^{t_1, t_2}$ is the \textbf{persistent relative homology} $H_k^{t_1, t_2}(S, A)$ and detects relative $k$-cycles in $S_{t_1}/A_{t_1}$ which are still present in $S_{t_2}/A_{t_2}$.
Given an open set $U \subset S_\infty$, define the \textbf{persistence module}
\begin{align}
\label{eq:persistent_relative_homology_module}
    \mathcal{H}^\bullet_k(U) &= \bigoplus_t H_k(S_t, S_t\backslash U) 
\end{align}
Then, our persistent homology feature spaces can be formally defined as the quotient
\begin{align}
\label{eq:persistent_relative_homology}
    \mathcal{H}_k(U) &= 
        \left( \bigoplus_t H_k(S_t, S_t\backslash U) \right) / \left( \bigoplus_{t_1 < t_2} \ima i_k^{t_1, t_2} \right) = 
        \mathcal{H}_k^\bullet(U) / \left( \bigoplus_{t_1 < t_2} \ima i_k^{t_1, t_2} \right)
\end{align}
The quotient removes the copies of a persistent cycle through its life interval.
Hence, the resulting space has a dimension for each unique persistent cycle.

\cite{sovdat2016text} studied a similar sequence where $A_t \subset S_t$ but not necessarily $A_t = S_t \cap A_\infty$ (i.e. a simplex can enter $S$ at a time step but also enter in $A$ at a later time step) and proposed an algorithm to compute this relative persistent (co)homology.

Apx.~\ref{subsec:computing_relative_homology} describes how persistent relative homology can be computed while Apx.~\ref{apx:sheaf_laplacian} describes a method to construct the corresponding sheaf Laplacian.


\section{Computing Relative Homology and Relative Persistent Homology}
\label{subsec:computing_relative_homology}

\paragraph{Computing Persistent Homology}
The \texttt{Ripser} library implements an efficient algorithm to compute persistent homology
 \cite{bauer2021ripser, ctralie2018ripser}. 
This algorithm can be easily adapted to also return the indices of the simplices which created and destroyed each homology class / persistent cycle; indeed, these indices are needed to implement a differentiable version of persistent homology \cite{bruel2019topology}.
Note that this software actually computes persistent co-homology and also returns representative cochains, which can be thought simply as the transpose of representative chains. 
In the rest of this section, we will work with \textbf{co-homology} groups $H^k(\cdot)$ rather than homology groups $H_k(\cdot)$ to better reflect the algorithm but we first emphasize that these groups are isomorphic.

Unfortunately, \texttt{Ripser} only compute absolute (co)homology.
\cite{sovdat2016text} previously described a very similar algorithm to compute the persistent \emph{relative} homology of a sequence of pairs $\{(S_t, A_t)\}_t$.
As discussed in Apx.~\ref{apx:persistent_homology}, they consider more general filterations than ours and, therefore, their algorithm is unnecessarily complicated for us.

Instead, we note that the Ripser algorithm from \cite{bauer2021ripser} essentially performs an (optimized) \emph{Gauss reduction} of the co-boundary matrix $\partial^\bullet_S: C^\bullet(S) \to C^\bullet(S)$, with rows and columns (corresponding to different simplices in the filtration) sorted by decreasing weight / birth time.
This algorithm can be used to compute the relative (co)homology $H^\bullet(S, A)$ by simply removing those rows and columns of $\partial^\bullet_S$ which belongs to $A$; indeed, by definition one obtains precisely the relative co-boundary map $\partial^\bullet_{S, A}: C^\bullet(S, A) \to C^\bullet(S, A)$ which defines relative (co)homology.

Moreover, as most existing persistent homology tools, \texttt{Ripser} only supports finite fields $\F = \Zp$ (for $p$ prime), while our sheaf requires features in the real field $\F = \R$.
Fortunately, the algorithm described in \cite{bauer2021ripser} works for any generic field $\F$, so \texttt{Ripser} can be easily adapted to compute (co)homology with $\F=\R$ coefficients.

\section{Computing the sheaf Laplacian}
\label{apx:sheaf_laplacian}

Let $A', B' \subset S$ be two open sets and $C' = A' \cap B' \subset S$ their intersection.
To construct the sheaf Laplacian between these two open sets $\Delta_{B',A'}^k = -\left[\gH^k(C' \subset A')\right]^T \circ \gH^k(C' \subset B')$ 
we need to construct the two restriction maps $\gH^k(C' \subset A'), \gH^k(C' \subset B')$ and then find equivalent cocycles in their images.

The following Mayer-Vietoris sequence for relative cohomology suggests a way to perform this computation.
Let $D' = A' \cup B'$ the union of the two open sets and $D = S \setminus D'$ its complementary; then the following sequence is exact:
\begin{align}
\label{eq:les_mayervietoris_relative_cohomology}
    \dots \to H^{k-1}(S, D) \to^{\partial^{k-1}} H^k(S, C) \to^{i_* \oplus -j_*} H^k(S, A) \oplus H^k(S, B) \to^{k_* + l_*} H^k(S, D) \to \dots
\end{align}
where the maps $i_*$ and $j_*$ are adjoint of the restriction maps $\gH^k(C' \subset A'), \gH^k(C' \subset B')$.
The co-boundary map $\partial^{k-1}$ detects the $k$ relative cycles in $C'$, not present in neither $A'$ nor $B'$, which have appeared when collapsing $C \setminus D$ in a single point (e.g. a line with it extremes in $D' \setminus C'$ is a connected component, i.e. a $0$-cycle, in $H^0(S, D)$, but when $D' \setminus C'$ is collapsed, the two extremes merge and the $0$-cycle becomes a $1$-cycle in $H^1(S, C)$).

This sequence implies that $H^k(S, C)$ splits as the co-image $\coima (i^* \oplus -j^*)$ (i.e. the image of the restriction maps) and the image $\ima \partial^{k-1}$.
In other words, the restrictions of two cocycles in $H^k(S, A)$ and $H^k(S, B)$ are equivalent if their difference is zero modulo $\ima \partial^{k-1}$.

Hence, we set up an \emph{extended coboundary matrix} $\gB^k$ whose reduction computes the sheaf Laplacian.

\textbf{Columns} The matrix columns are divided in two sets.
First, it contains all columns of $\partial^\bullet(S, D)$ as used in Apx.~\ref{subsec:computing_relative_homology} to compute the persistent relative cohomology $H^k(S, D)$.
Second, it contains a column for each persistent cocycle found previously in $H^k(S, A)$ and $H^k(S, B)$.
Like in Apx.~\ref{subsec:computing_relative_homology}, the columns in the first set are sorted inversely by the weight of each $k-1$ simplex in $D'$.
Instead, the columns in the second set are sorted inversely by their corresponding cocycle's birth time (cocycles of $A'$ and $B'$ are mixed by sorting).
These two sets split the matrix in two sub-matrices $\gB^k = [\gB^k_D, \gB^k_{AB}]$.

\textbf{Rows in $\gB^k_D$}
Columns in $\gB^k_D$ simply contain the coboundaries in $D'$ of each simplex, sorted by decreasing weight, as in Apx.~\ref{subsec:computing_relative_homology}.

Before defining the rows in $\gB^k_{AB}$,
let's first recall some details about the algorithm in \cite{bauer2021ripser}.
A cocycle in $H^k(S, A)$ (or $B$) can be represented by the column of the reduction matrix used to reduce $\partial^k_A$.
This vector expresses a $k$-cocycle as a linear combination of $k$-simplices in $A'$.
The non-zero simplex with lowest weight defines the birth time of the cocycle.
The corresponding reduced column contains the coboundary and the first non-zero $k+1$ simplex (the \emph{pivot}) defines the death time of the cocycle (since, after that time, the cocycle doesn't belong to the kernel of the coboundary map anymore).

\textbf{Rows in $\gB^k_{AB}$}
The columns in $\gB^k_{AB}$ contain three sets of row.
Each column, corresponding to a certain cocycle to restrict, has
\begin{enumerate}
	\item one row for each $k$-simplex in $D'$: these rows contain a copy of the reduction vector representing the cocycle as above (note that $A', B' \subset D'$). These are also the same rows in $\gB^k_D$
	\item one row for each $k+1$ simplex in $A'$: these rows contain a copy of the coboundary of the cocycles in $A'$
	\item another row for each $k+1$ simplex in $B'$: these rows contain a copy of the coboundary of the cocycles in $B'$
\end{enumerate}
Note that each $k+1$ simplex in $D$' appears twice in the rows.

A linear combination of the columns of this extended reduction matrix is a linear combination of cocycles in $H^k(S, A)$, $H^k(S, B)$ and $H^{k-1}(S, D)$.
This represents a pair of cocycles $\gamma_A \in H^k(S, A)$ and $\gamma_B \in H^k(S, B)$ and the rows in the resulting column model the three constraints we are trying to enforce.
Indeed, a non-zero value in a row implies
\begin{itemize}
	\item if the row is a $k+1$-simplex in $A'$ (or $B'$), the cocycle $\gamma_A \in H^k(S, A)$ (or $\gamma_B \in H^k(S, B)$) is dead at this time step (and so must be also its restriction to $H^k(S, C)$ as proved in Theorem~\ref{thm:restriction_dies_earlier}).
	\item if the row is a $k$-simplex in $C' \subset D'$, it means that the sum of $\gamma_A$ and $\gamma_B$ is not zero at this time step i.e. their restrictions are not equivalent cocycles.
	\item if the row is a $k$-simplex in $D' \setminus C' = (A' \setminus C') \cup (B' \setminus C')$, either $\gamma_A$ or $\gamma_B$ can not be restricted to $H^k(S, C)$ at this time step.
\end{itemize}

Then, the matrix reduction algorithm trying to find pairs of cocycles which satisfy these constraints for the longest time.
Once this matrix is reduced, a column in $\gB^k_{AB}$ represents a pair of cocycles $\gamma_A \in H^k(S, A)$ and $\gamma_B \in H^k(S, B)$ whose sum is $0$ when restricted to $H^k(S, C)$, modulo the coboundary of some cocycles in $H^{k-1}(S, D)$, until the time step the \emph{pivot} of this column appears in the filtration.
Then, the pivot corresponds to the time step one of the three constraints above is violated.

Hence, the reduced columns in $\gB^k_{AB}$ can be used to construct the sheaf Laplacian as follows.
Let the $i$-th reduced column correspond to a pair of cocycles $(\gamma_A^i, \gamma_B^i)$ which are obtained by linearly combining the persistent bases of $H^k(S, A)$ and $H^k(S, B)$ via the reduction vectors $\vv_A^i$ and $\vv_B^i$, respectively.
Note that these reduction vectors essentially construct the two restriction maps.
Let $t_i$ be the time the pivot of this column appear and let $s_A^i$ be the birth time of $\gamma_A^i$ (i.e. the lowest weight of its simplices) and $t_A^i$ its death time, and $s_B^i$ and $t_B^i$ those of $\gamma_B^i$.
This pair restricts to the same cocycle in $H^k(S, C)$ only in the time interval $[s^i, t^i)$, with $s_i = \max(s_A^i, s_B^i)$ and $t_i \leq s_B^i, s_A^i$ due to Theorem~\ref{thm:restriction_dies_earlier}.
The pair $(s^i, t^i)$ defines the time interval during which an $i$-th sheaf Laplacian persists:
\[
	[\Delta_{\gH^k}^i]_{A', B'} = \vv_A^i (\vv_B^i)^T
\]
We do not include the $-1$ sign since our constraint enforced $\gamma_A + \gamma_B \cong 0$, i.e. $\gamma_A \cong - \gamma_B$.
This Laplacian is visualized also in Fig.~\ref{fig:persistent_laplacian}.

Note that the non-zero coefficients in the vector $\vv_A^i$ or $\vv_B^i$ are associated with persistent cocycles of $H^k(S, A)$ or $H^k(S, B)$ which might appear and die at different time steps.
It follows that each entry of $[\Delta_{\gH^k}^i]_{A', B'}$ has an independent persistence interval given by the intersection of $[s^i, t^i)$ with the intervals of the two cocycles of $A'$ and $B'$ involved.

If we define $\vv|_t$ as the components of $\vv$ which are "active" at time $t$, the sheaf Laplacian at a time step $t$ can be constructed as 
\[
	[\Delta_{\gH^k}^t]_{A', B'} = \sum_{i : t \in [t^i, s^i)} \vv_A^i|_t (\vv_B^i|_t)^T
\]

Finally, the embedding and projection operations mention in Sec.~\ref{sec:method} can be easily implemented by weighting the entry $(a, b)$ of the matrix $[\Delta_{\gH^k}^i]_{A', B'}$ by its own life span $\min(t_i, t_a, t_b) - \max(s_i, s_a, s_b)$ divided by the output cocycle life span $t_a - s_a$.

\subsection{Other properties of the Local (Co)Homology Sheaf}

The following properties guarantee the intuitive fact that (co)cycles appear and disappear first in smaller neighborhoods than in larger ones.
In other words, if a (co)cycles is in the image of the restriction map $\gH^k(A^t \subset B^t)$ at time $t$, then it also needs to be in the image at any previous time steps (until the birth time in $\gH^k(B)$); similarly, if a (co)cycles is in the kernel of $\gH^k(A^t \subset B^t)$ at a time step $t$, it will also be at any following time steps (until its death in $\gH^k(B)$).

\begin{theorem}[The restriction of a cocycle dies earlier]
\label{thm:restriction_dies_earlier}
Consider the following \emph{commutative} diagram for relative persistent cohomology and assume a single simplex is added to $\mathbb{S}$ at each time step $t$:
\begin{equation}
\label{eq:relative_cohomology_les}
\begin{tikzcd}
	{\cdots } & {H^{k-1}(S_t)} & {H^{k-1}(A_t)} & {H^{k}(S_t, A_t)} & {H^k(S_t)} & \cdots \\
	\\
	\cdots & {H^{k-1}(S_{t+1}) } & {H^{k-1}(A_{t+1})} & {H^k(S_{t+1}, A_{t+1})} & {H^k(S_{t+1}) } & \cdots
	\arrow["{\partial^{k-1}_t}", from=1-3, to=1-4]
	\arrow["{i_t^*}", from=1-4, to=1-5]
	\arrow["{j^*_t}", from=1-2, to=1-3]
	\arrow["{\partial^{k-1}_{t+1}}", from=3-3, to=3-4]
	\arrow["{i_{t+1}^*}", from=3-4, to=3-5]
	\arrow["{j^*_{t+1}}", from=3-2, to=3-3]
	\arrow["{f^{t, t+1}_{A}}"{description}, from=1-3, to=3-3]
	\arrow["{f^{t, t+1}_{S, A}}"{description}, from=1-4, to=3-4]
	\arrow["{f^{t, t+1}_{S}}"{description}, from=1-5, to=3-5]
	\arrow["{f^{t, t+1}_{S}}"{description}, from=1-2, to=3-2]
	\arrow[from=1-5, to=1-6]
	\arrow[from=3-5, to=3-6]
	\arrow[from=1-1, to=1-2]
	\arrow[from=3-1, to=3-2]
\end{tikzcd}
\end{equation}
Let $\gamma \in \ima{i^*_t} \subset H^k(S_t)$ be a cocycle of $S_t$ at time $t$ which corresponds to a relative cocycle $\bar{\gamma} \in H^k(S_t, A_t)$, i.e. $\gamma = i_t^*(\bar{\gamma})$.
Assume that at time $t+1$ a $k+1$-simplex $\sigma$ is added to $S_t$ such that the cocycle $\gamma$ dies in $H^k(S_{t+1})$, i.e. $\gamma \notin \ima{f^{t,t+1}_S}$.
Then, $\bar{\gamma} \notin \ima{f^{t,t+1}_{S, A}}$ either and, therefore, the relative cocycle $\bar{\gamma}$ dies at time $t+1$, too.
\begin{proof}
   Let $\gamma \in \ker{f^{t, t+1}_S}$ and let $\sigma$ be the $k+1$ simplex added in $S_{t+1}$ which killed $\gamma$ (i.e. $S_{t+1} = S_t \cup \{\sigma\}$).
   Assume $\exists \bar{\gamma} \in H^k(S_t, A_t)$ such that $\gamma = i_t^*(\bar{\gamma})$.
   
   Since $\sigma$ is a $k+1$-simplex, $H^{k-1}(A_{t+1}) \cong H^{k-1}(A_t)$ and $H^{k-1}(S_{t+1}) \cong H^{k-1}(S_t)$.
   Because these cohomology groups did not change, $\ima{j^*_{t}} \cong \ima{j^*_{t+1}}$ and, therefore, $\ker{\partial^{k-1}_{t}} \cong \ker{\partial^{k-1}_{t+1}}$.
   It also follows that $\coima{\partial^{k-1}_{t}} \cong \coima{\partial^{k-1}_{t+1}}$, i.e. $\ker{i^*_t} = \ker{i^*_{t+1}}$.
   
   Finally, because $\bar{\gamma} \notin \ker{i^*_t} \cong \ker{i^*_{t+1}}$,
   $\bar{\gamma} \in H^{k+1}(S_{t+1}, A_{t+1}) \cong \ker{i^*_{t+1}} \oplus \coima{i^*_{t+1}}$ if and only if 
   $\bar{\gamma} \in \coima{i^*_{t+1}}$.
   This requires that $\exists \gamma' = i^*_{t+1}(\bar{\gamma}) \in H^k(S_{t+1})$.
   However, the commutativity of the diagram guarantees that 
   $i^*_{t+1}(f_{S, A}^{t, t+1}(\bar{\gamma})) = f_{S}^{t, t+1}(i^*_{t+1}(\bar{\gamma})) = 0$, which is a contradiction.
   Hence, $\bar{\gamma} \in \ker{f^{t, t+1}_{S, A}}$, i.e. the relative cocycle $\bar{\gamma}$ must also die at time $t+1$.
\end{proof}
\end{theorem}
A similar argument should work also for triples, i.e. projections $H^k(S, B) \to H^k(S, A)$ with $B \subset A \subset S$ by replacing $H^k(S)$ with $H^k(S, B)$ and $H^k(A)$ with $H^k(A, B)$.

\begin{theorem}[The restriction of a cocycle appears earlier]
\label{thm:restriction_appears_earlier}
Consider again the \emph{commutative} diagram for relative persistent cohomology in Eq.~\ref{eq:relative_cohomology_les} (here, shifted right by two steps):
\begin{equation}
\begin{tikzcd}
	\cdots & {H^{k}(S_t, A_t)} & {H^k(S_t)} & {H^k(A_t)} & {H^{k+1}(S_t, A_t)} & \cdots \\
	\\
	\cdots & {H^k(S_{t+1}, A_{t+1})} & {H^k(S_{t+1}) } & {H^k(A_{t+1})} & {H^{k+1}(S_{t+1}, A_{t+1})} & \cdots
	\arrow["{i_t^*}", from=1-2, to=1-3]
	\arrow["{i_{t+1}^*}", from=3-2, to=3-3]
	\arrow["{f^{t, t+1}_{S, A}}"{description}, from=1-2, to=3-2]
	\arrow["{f^{t, t+1}_{S}}"{description}, from=1-3, to=3-3]
	\arrow["{j^*_t}", from=1-3, to=1-4]
	\arrow["{\partial^k_t}", from=1-4, to=1-5]
	\arrow["{j^*_{t+1}}", from=3-3, to=3-4]
	\arrow["{\partial^k_{t+1}}", from=3-4, to=3-5]
	\arrow["{f^{t, t+1}_{A}}"{description}, from=1-4, to=3-4]
	\arrow[from=1-5, to=1-6]
	\arrow[from=3-5, to=3-6]
	\arrow["{f_{S,A}^{t, t+1}}"{description}, from=1-5, to=3-5]
	\arrow[from=1-1, to=1-2]
	\arrow[from=3-1, to=3-2]
\end{tikzcd}
\end{equation}

Again, assume a single simplex is added to $\mathbb{S}$ at each time step $t$.
Let $\gamma \in H^k(S_{t})$ be a cocycle of $S_{t}$ which persists to $S_{t+1}$, i.e. $\gamma \in \ima{f_S^{t,t+1}}$.

Assume that there exists a relative cocycle $\bar{\gamma} \in H^k(S_{t+1}, A_{t+1})$ such that $\gamma = i_{t+1}^*(\bar{\gamma})$.

Then, $\bar{\gamma} \in \ima{f^{t,t+1}_{S, A}}$, too.
This implies that the projection $\bar{\gamma}$  must always appear in the filtration at the same time or earlier than the corresponding cocycle $\gamma = i^*(\bar{\gamma})$.

\begin{proof}

    Assume $\gamma \notin \ima{i^*_{t}}$.
    Then, there exists a new relative cocycle $\bar{\gamma}'$ in $H^k(S_{t+1}, A_{t+1})$  appearing at time $t+1$, with $\gamma = i^*_{t+1}(\bar{\gamma}')$.
    Let $\sigma$ be the $k$-simplex added to $S_{t} \setminus A_t$ which gave birth to it  (i.e. $S_{t+1} = S_t \cup \{\sigma\}$ and $A_t = A_{t+1}$). 
    Since $\sigma \notin A_{t+1}$, $H^*(A_{t+1}) \cong H^*(A_t)$.
    Moreover, since $\sigma$ is a $k$-simplex, $H^{k+1}(S_t, A_t) \cong H^{k+1}(S_{t+1}, A_{t+1})$, too.

    It follows that $\ker{\partial^k_t} \cong \ker{\partial^k_{t+1}}$ and, therefore, $\coima{j^*_{t+1}} \cong \coima{j^*_{t}}$.
    Since $\gamma \in \ima{i^*_{t+1}}$, $\gamma \notin \coima{j^*_{t+1}} \cong \coima{j^*_{t}}$.
    Hence, $\gamma \in \ker{j_t^*} \cong \ima{i^*_t}$.
    
    This a contradiction, so it must be the case that $\gamma \in \ima{i^*_t}$, too.

    Now, let $\bar{\gamma} \in H^k(S_t, A_t)$ s.t. $\gamma = i_t^*(\bar{\gamma})$.
    Then, by commutativity of the diagram, $\gamma = f_S^{t,t+1}(i_t^*(\bar{\gamma})) = i^*_{t+1}(f_{S,A}^{t,t+1}(\bar{\gamma}))$, which implies $\bar{\gamma} \in \coima{f_{S,A}^{t,t+1}}$.
    In other words, $\bar{\gamma}$ is also a persistent cocycle in $H^k(S, A)$.
    
\end{proof}
\end{theorem}

As earlier, a similar argument should work also for triples $B \subset A \subset S$.

\end{document}